\pdfoutput=1
\documentclass[11pt]{article}

\usepackage[final]{ACL2023}

\usepackage{times}
\usepackage{latexsym}

\usepackage[T1]{fontenc}
\usepackage[utf8]{inputenc}

\usepackage{microtype}

\usepackage{inconsolata}

\usepackage{tabularx}
\usepackage{booktabs}
\usepackage{boldline}
\usepackage{subcaption}
\usepackage{multirow}
\usepackage{float}
\usepackage{xcolor}
\usepackage{graphicx}
\usepackage{dialogue}
\usepackage{csquotes}
\usepackage{listings}
\usepackage{colortbl}
\usepackage{color}

\newcommand\blfootnote[1]{%
  \begingroup
  \renewcommand\thefootnote{}\footnote{#1}%
  \addtocounter{footnote}{-1}%
  \endgroup
}

\title{{Revealing User Familiarity Bias in Task-Oriented Dialogue \\ via Interactive Evaluation}}

\author{
    Takyoung Kim$^{1,\dagger}$ \: Jamin Shin$^{1,2,\ddagger}$ \: Young-Ho Kim$^{1}$ \: Sanghwan Bae$^{2}$ \: Sungdong Kim$^{1,2}$ \\ \\
    NAVER AI Lab$^{1}$ \; NAVER Cloud$^{2}$\\
    \href{mailto:tk30@illinois.edu}{\texttt{tk30@illinois.edu}} \quad \href{mailto:jayshin.nlp@gmail.com}{\texttt{jayshin.nlp@gmail.com}} \quad \href{mailto:yghokim@younghokim.net}{\texttt{yghokim@younghokim.net}} \\ 
    \texttt{\{sanghwan.bae, sungdong.kim\}@navercorp.com} 
}

\begin{document}
\maketitle

\blfootnote{\textsuperscript{$\dagger$}Work done as a research intern at NAVER AI Lab.}
\blfootnote{\textsuperscript{$\ddagger$}Work done while at NAVER AI Lab.}

\newcommand{\todoc}[2]{{\textcolor{#1}{\textbf{#2}}}}
\newcommand{\todoblue}[1]{\todoc{blue}{\textbf{#1}}}
\newcommand{\todored}[1]{\todoc{red}{#1}}
\newcommand{\todobrown}[1]{\todoc{brown}{\textbf{#1}}}
\newcommand{\todocyan}[1]{\todoc{cyan}{\textbf{#1}}}
\newcommand{\todopurple}[1]{\todoc{purple}{\textbf{#1}}}

\newcommand{\sung}[1]{\todored{sung: #1}}
\newcommand{\tak}[1]{\todobrown{tak: #1}}
\newcommand{\jay}[1]{\todoblue{jay: #1}}
\newcommand{\bay}[1]{\todocyan{bay: #1}}
\newcommand{\ho}[1]{\todopurple{ho: #1}}

\newcommand{\ie}{{\it i.e.}}%
\newcommand{\eg}{{\it e.g.}}%

\definecolor{tableheader}{HTML}{EFEFEF}

\newcommand{\symbolbot}{\raisebox{-2pt}{\includegraphics[width=11pt]{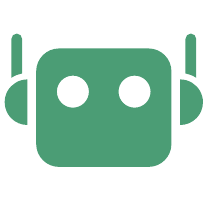}} }

\newcommand{\symboluser}{\raisebox{-1.7pt}{\includegraphics[width=10pt]{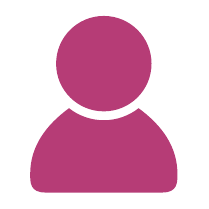}} }

\newcommand{\userdial}[1]{{\fontsize{10}{11}\textit{#1}}\vspace{1mm}}
\newcommand{\botdial}[1]{{\fontsize{9.5}{9}\texttt{#1}}\vspace{1mm}}

\newcommand{\update}[1]{\textcolor{black}{#1}}

\begin{abstract}
Most task-oriented dialogue (TOD) benchmarks assume users that know exactly how to use the system by constraining the user behaviors within the system's capabilities via strict user goals, namely ``user familiarity'' bias.
This data bias deepens when it combines with data-driven TOD systems, as it is impossible to fathom the effect of it with existing static evaluations.
Hence, we conduct an interactive user study to unveil how vulnerable TOD systems are against realistic scenarios.
In particular, we compare users with 1) detailed goal instructions that conform to the system boundaries (\textit{closed-goal}) and 2) vague goal instructions that are often unsupported but realistic (\textit{open-goal}).
Our study reveals that conversations in open-goal settings lead to catastrophic failures of the system, in which 92\% of the dialogues had significant issues.
Moreover, we conduct a thorough analysis to identify distinctive features between the two settings through error annotation. 
From this, we discover a novel ``pretending'' behavior, in which the system pretends to handle the user requests even though they are beyond the system's capabilities.
We discuss its characteristics and toxicity while \update{showing recent large language models can also suffer from this behavior}.
\end{abstract}

\section{Introduction}
\label{intro}

Task-oriented dialogue (TOD) systems aim to accomplish specific user goals by comprehending their requests and making appropriate API calls or database (DB) searches~\cite{young-2013-pomdp}. 
TOD systems typically use a pipeline approach, connecting separate modules such as intent detection, dialogue state tracking, policy management, and natural language generation, often requiring complex rules or heuristics.
End-to-end (E2E) TOD systems have been highlighted as a fully data-driven solution because of their concise implementation~\citep{bordes2017learning, wen-etal-2017-network}. Recently, such TOD systems have significantly improved on top of pre-trained language models~\cite{asl-etal-2020-simpletod, ham-etal-2020-end, peng-etal-2021-soloist, he2022galaxy}.
\begin{figure}[t!]
    \centering
    \includegraphics[width=\columnwidth]{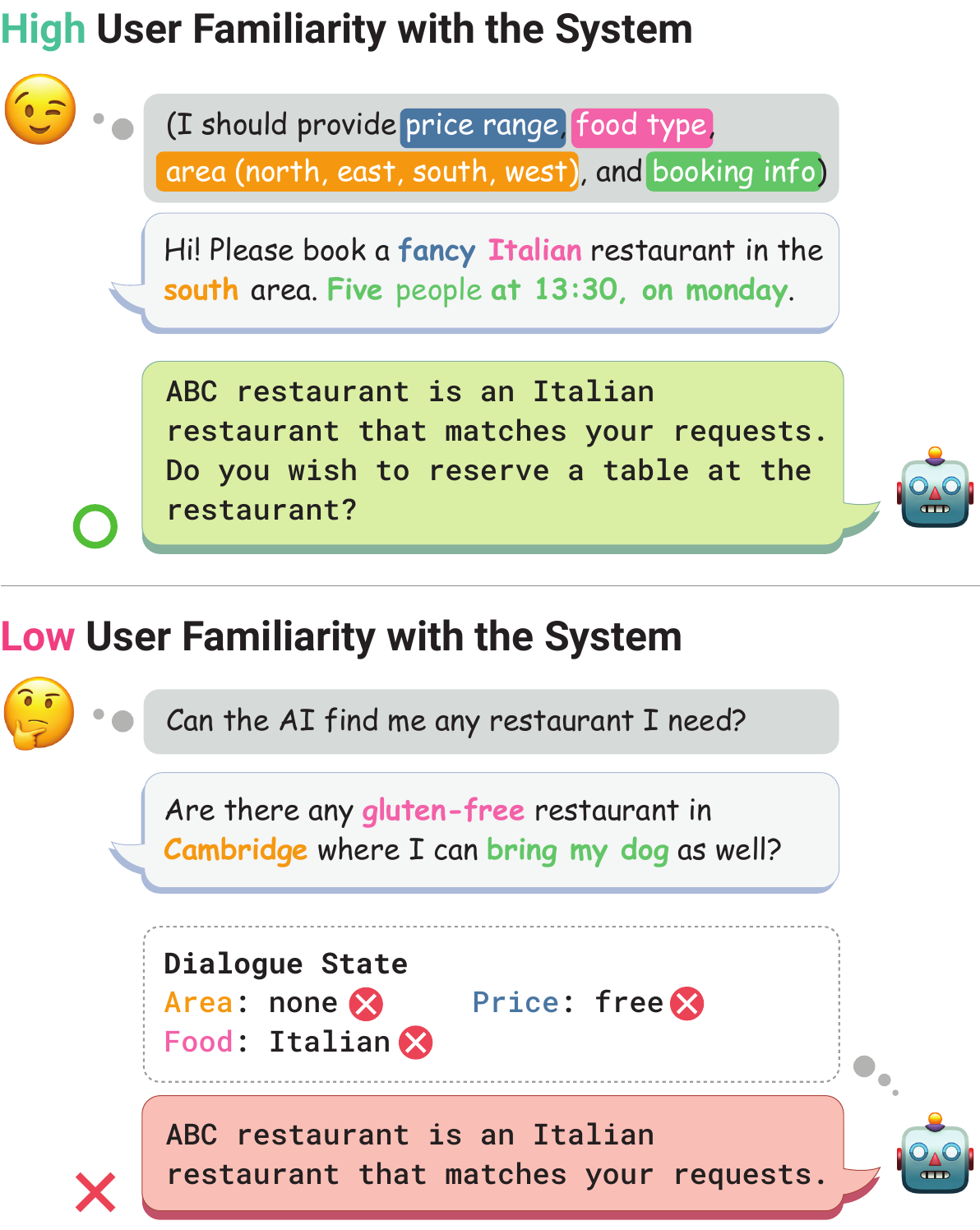}
    \caption{Contrastive dialogues according to user familiarity with the system. Users with high familiarity converse with the system within the predefined scenario since they already know the system well. However, users with low familiarity are more likely to talk about a broad range of topics beyond the system's capacities leading to the erroneous response of the system.}
    \label{fig:example}
    \vspace{-3mm}
\end{figure}

However, despite the numerous studies on TOD systems and the great successes of large language models, we argue that there is a huge gap between the TOD studies and deployable TOD systems. 
Among the many reasons hindering end-to-end systems from being widely adopted in the industry, the instability of such systems makes it harder to match the robustness of conventional pipeline systems.

We hypothesize that the major source of this instability lies in the naive assumption about the users during TOD data collection. 
We call this the \textit{user familiarity} bias, as illustrated in Figure~\ref{fig:example}.
For instance, during Wizard-of-Oz style data collection~\cite{kelley1984iterative}, the user-role workers are provided with detailed instructions on the goal they need to achieve which conforms with the system capabilities~\cite{budzianowski-etal-2018-multiwoz, byrne-etal-2019-taskmaster}.
Hence, as the user behaviors are strictly constrained, this process simulates users who know exactly how to use the system~\cite{larson-etal-2019-evaluation}.
Other datasets based on user simulation, such as M2M and SGD~\cite{shah-etal-2018-bootstrapping, rastogi2020towards}, include the same user familiarity bias, as they simulate users based on predefined user goals and rules specifying how to converse.
On the other hand, real users in the wild often have fairly creative or vague goals way beyond the system coverage, and this user-side bias prevents us from modeling such realistic conversations.

In this paper, we conduct an interactive user study on one of the most-used Wizard-of-Oz-styled TOD benchmarks, MultiWOZ~\cite{budzianowski-etal-2018-multiwoz}, to investigate the impact of \textit{user familiarity}.
The main objective of this study is to determine whether the familiarity of the user with the TOD system's capabilities influences the successful completion of a conversational task.
To do this, we first divide the users into two groups: \textit{closed-goal} and \textit{open-goal}.
The former user group is provided with detailed user goal instructions that are from the MultiWOZ, while the latter is given only a portion of the instructions along with some realistic goals that are not supported by the system, thereby simulating users who are not familiar with TOD systems.
Based on each goal type they are assigned to, the users converse with a recent E2E TOD system, GALAXY~\cite{he2022galaxy}, \update{which is trained on diverse TOD datasets including MultiWOZ}.

Our user study reveals that 92\% of the dialogues in the open-goal setting has significant issues that often lead to failure in achieving the user goals. 
Moreover, we find that various inconveniences caused by the TOD system force users to quit the conversation regardless of the goal types.
We thoroughly analyze the resulting conversations to identify the impact of user familiarity by annotating erroneous turns. In particular, we figure out six prevalent errors in both goal settings. As expected, open-goal dialogues contain more problematic turns, and the open-goal setup causes more irrelevant and unfaithful responses.

In addition, we identify unfaithful responses as ``pretending'' behaviors of the system that primarily arises in the open-goal setting. 
This is a phenomenon similar to hallucination, in which the dialogue system pretends to handle the user's requests even though they are beyond the system's boundaries, but more potentially harmful because it is \update{\textbf{almost impossible}} for users to verify the reliability of the information during conversation since the hallucinated pieces of information are usually \textit{service-specific}.
We believe this issue is relatively underexplored as we witness most previous works focused on the closed-goal setting, and our qualitative analysis of the open-goal dialogues demonstrates that such pretending behaviors are prevalent and crucial.

\update{Finally, we conduct case studies to check whether recent large language models with strong zero-shot performance can mitigate each conversational error. We show that large language models are proficient to handle errors within given context, but preventing pretending problem highly depends on the system design, not only on language models' performance.}

Our contributions are threefold: (1) interactive user study that breaks away from the closed-goal assumption; (2) examination of the characteristics of erroneous situations in both dialogue- and turn-levels; and (3) demonstration of the ``pretending'' problem of the TOD systems, especially as observed in an open-goal assumption, where the agent deceives users as if it handled their exceptional requests.

\section{Related Works}

\subsection{TOD Benchmarks}
\label{rw:benchmark}
MultiWOZ is one of the largest TOD benchmarks containing about 10,000 multi-domain dialogues~\cite{budzianowski-etal-2018-multiwoz}, following the Wizard-of-Oz protocol~\cite{kelley1984iterative}. In this setup, human workers participate in a conversation as either a user or system role. To guide the user-side workers to behave within the desired scenarios, the authors provide \textit{goal instructions} to the user-side worker detailing what to inform and request to the system-side worker. Meanwhile, the Taskmaster-1 dataset is more severe in that each conversation of the dataset is collected by one human worker playing both user and system roles, namely the self-play method~\cite{byrne-etal-2019-taskmaster}. Thus, these datasets naturally contain significant \textit{user familiarity} bias. Similarly, other datasets constructed by an automatic user simulator also contain the same bias since the simulation is based on predefined goals and rules \textit{bound} to the system's coverage~\cite{shah-etal-2018-bootstrapping, rastogi2020towards}.

\subsection{Benchmark Reality in TOD}
Recently, there have been studies concerning the reality of the benchmark dataset in the area of TOD. \citet{kim-etal-2020-beyond} incorporate an external unstructured knowledge (\ie, FAQ) to complement the language model trained on limited scenarios. Even though the dataset includes knowledge-seeking turn detection to handle out-of-scope requests, it still assumes high user familiarity with the system in that users require information specified in the external knowledge. \citet{qian-etal-2022-database, kim-et-al-2022-ohmymistake, yang-et-al-2022-toward} point out the limited coverage of dialogue expression by modifying the utterances of the user and system. Furthermore, \citet{sun-etal-2021-adding, li-et-al-2022-enhancing, young2022fusing} improve the model's natural conversation skills in terms of engagingness by combining with open-domain dialogue. However, we believe the combination cannot be a solution for dealing with users who have open-ended goals. On the other hand, \citet{qin-etal-2021-dont} argue inconsistent responses can be a more critical problem in TOD, and propose an accompanying new dataset to mitigate it.

\subsection{\update{TOD in Deployment}}
\update{Potential issues related to interaction or deployment were discussed among communities. For example, \citet{larsson-2017-user} mainly discussed technical and architectural difficulties in deploying dialogue systems. More similarly, \citet{leuski-artstein-2017-lessons} presented challenges where dialogue systems do not properly handle users' sub-dialogues of different topics or domains (\ie, lack of affordance). However, these discussions were mainly tested on proprietary products such as Siri and Alexa.}

\subsection{Evaluation of TOD System}
Many recent works evaluate performance using quantitative metrics for predefined slots and responses. Specifically, \citet{budzianowski-etal-2018-multiwoz} define two task-specific metrics, Inform and Success rate, which measure how often the system has provided the appropriate entities and answered all the requested attributes. In addition, BLEU~\cite{papineni-etal-2002-bleu} is used to measure the fluency of the generated responses.
However, \citet{nekvinda-dusek-2021-shades} report inconsistencies in data preprocessing for these metrics in an attempt to make standardized evaluation scripts.
Furthermore, \citet{cheng-etal-2022-simulator} build a user simulator capable of dynamic evaluation to solve the static evaluation problem for TOD systems. However, the evaluation is still limited to the closed-goal setup.

Apart from the automatic quantitative evaluation, there are consistent works of user evaluation in spoken dialogue research \citep{10.1145/274644.274722, ai-etal-2007-comparing, gasic-etal-2008-training}. Our work is more closely inspired by user studies in the Human-Computer Interaction (HCI) area that investigated live interactions between chatbots and users.
In particular, \citet{yeh-et-al-2022-guide} investigate the impacts of various guidance types and timing on performance metrics for TOD systems.
\citet{li-etal-2020-conversationanalysis} analyze conversation logs between users and TOD chatbots and claimed that identifying conversational ``non-progress'' is crucial for improving chatbot usability.
\section{Interactive User Study}

In this section, we explain the experimental setups of our interactive user studies on the current state-of-the-art data-driven TOD model.
Our focus lies on creating realistic scenarios, breaking away from evaluation solely based on TOD benchmarks. 
In particular, we are curious about the influence of user familiarity on the TOD system. 
We describe the details of the study in the following sections.

\subsection{User Goal}

Most TOD systems assume the users have specific goals in a given domain, \eg, restaurant or hotel reservations. 
Typically, such goals can be represented by sentences in natural language to control user-side human participants when collecting dialogue data~\cite{budzianowski-etal-2018-multiwoz, byrne-etal-2019-taskmaster}. 
The following is one of the user goal instructions provided in \citet{budzianowski-etal-2018-multiwoz}.
\begingroup
\addtolength\leftmargini{-0.2in}
\begin{quote}
\fontsize{10}{10}\textit{You are looking for a \textbf{place to stay}. The hotel should be in the \textbf{north} and should \textbf{include free parking}. The hotel should be in the type of \textbf{guesthouse}. Once you find the hotel, you want to book it for \textbf{3 people} and \textbf{2 nights} starting from \textbf{wednesday}. ... Make sure you get the \textbf{reference number}.}
\end{quote}
\endgroup

However, all user goals in most TOD benchmarks are based on a naive assumption that the users have sufficient knowledge about the dialogue system in advance. 
Thus, conversations based on such goals are always within expected scenarios from the TOD system's point of view.  
On the other hand, we argue that most real users are not familiar with TOD systems, and such users are prone to making exceptional requests beyond the system's capacity. 
To investigate the impact of user familiarity, we set up two user groups that have different types of goals considering their familiarity with TOD systems, which we refer to as closed-goal and open-goal, respectively.

\paragraph{Closed Goal}
Closed goals contain predefined scenarios which TOD systems can accomplish easily. 
In other words, it does not include any exceptional requests or actions from the perspective of the tested TOD system.
As we mentioned, most dialogues in existing TOD datasets constructed based on such predefined user goals fall within the capacity that the system can correspond to. 
We use these user goals from the \textit{restaurant} and \textit{hotel} domains of MultiWOZ~\cite{budzianowski-etal-2018-multiwoz} as our set of closed goals.

\paragraph{Open Goal}

Contrary to the closed-goal setting, open-goal settings are used to simulate realistic situations for users who have little idea about the TOD systems except for the domain. 
Real-world users may have a wider range of purposes than the predefined situations because the system capacity cannot include every possible scenario within its boundaries. 
Thus, we include exceptional requests which are not covered by the original dataset. 
Specifically, we create an open-goal by inserting the exceptional requests into a subset of closed-goal.
By doing so, we are able to cover essential user requirements covered by the system (\eg, time to visit), while also simulating real-life requests that are unsupported. 
In our experiments, we limit the number of exceptional requests in a single open-goal to a maximum of two.

To construct the set of exceptional requests, we use InstructGPT~\cite{brown-etal-gpt3, ouyang2022training} \texttt{text-davinci-003} by OpenAI API. 
First, we input high-level task information as a prompt and let InstructGPT come up with the remaining requests to complete the task instruction. 
Table~\ref{table:prompt} is an example input prompt and output of the generated exceptional request. 
Then, we manually validate the generated outputs to filter out noisy outputs and categorize commonly observed requests for the target domains into an ontology, which is shown in Appendix Listing~\ref{list:ontology}.

\begin{table}[t]
    % Original arraystretch: 1.4
    \centering\small\ttfamily
			\def\arraystretch{1.5}
		    \setlength{\tabcolsep}{0.5em}
    \begin{tabularx}{\columnwidth}{|>{\raggedright}X|}
     \hline
     \rowcolor{tableheader}\textbf{\textsf{Input Prompt}} \tabularnewline
     \arrayrulecolor{lightgray}\hline
     \small
     Imagine that you are planning to travel UK. The following goal is user specification to
     find information from the bot. Freely fill 
     in the remaining specification. \\
     \small
     (Goal) You are looking for a \textbf{place to stay}. \tabularnewline
     \arrayrulecolor{black}\hline
     \rowcolor{tableheader}\textbf{\textsf{Example Output}} \tabularnewline
     \arrayrulecolor{lightgray}\hline
     \small
     Find a hotel that is \textbf{nearby Cambridge city}, 
     \textbf{close to public transportation, good}
     \textbf{customer reviews} from past guests, include
     \textbf{daily meals} in the cost, \textbf{WiFi} included, and
     \textbf{reasonable cost}. \tabularnewline
     \arrayrulecolor{black}\hline
    \end{tabularx}
    \caption{An example input and output of exceptional requests generation based on InstructGPT. We guide InstructGPT to perform text completion from the given input prompt, and manually select exceptional requests not handled in MultiWOZ. After collecting generated outputs, we construct an open-goal ontology demonstrated in the Appendix Listing \ref{list:ontology}.}
    \label{table:prompt}
    \vspace{-3mm}
\end{table}

\subsection{Participants}
We recruit 20 participants working at a tech company who meet our inclusion criteria\footnote{Refer to Appendix~\ref{appendix:participant} for exhaustive demographics.}
: (1) having some experience with AI chatbots and (2) feeling comfortable carrying on written conversations in English. In terms of the number of participants, our sample size is congruent with the guidelines and standards for the sample size for HCI studies \citep{10.1145/1735223.1735255, 10.1145/2858036.2858498}. Following suggestions by \citet{10.1145/1735223.1735255}, we aim to recruit around 10 participants per group. 
We randomly assign participants one of the two conditions: Open-goal ($N=10$; referred to as \texttt{O1}--\texttt{O10}; 7 females) or Closed-goal ($N=10$; referred to as \texttt{C1}--\texttt{C10}; 5 females). 
Each group of participants is provided with the corresponding type of goal instructions. 

\subsection{Procedure}
We implement a chat interface on Gradio~\citep{abid2019gradio} web platform illustrated in Appendix Figure~\ref{fig:demo}. 
For the system agent, we use one of the most performant\footnote{Based on the official MultiWOZ leaderboard: \url{https://github.com/budzianowski/multiwoz}.} E2E TOD model~\citep{he2022galaxy} trained on diverse benchmarks including MultiWOZ 2.1~\citep{eric-etal-2020-multiwoz}. Please note that this TOD model also contains the user familiarity bias in TOD benchmark datasets, and our experiment can be generalized to every model trained with datasets constructed with similar manner (refer to Section~\ref{rw:benchmark}).
We give users structured goals instead of the sentence format in order to prevent copying biases brought on by sentences. 
After completing each conversation, participants are instructed to specify whether they finish the conversation until the end (whether the goal is achieved) and whether they have encountered any significant inconveniences. 

If participants mention that they cannot properly complete the conversation or they experience any inconvenience, we prompt a follow-up checkbox field to ask the categories of inconveniences: 
(1) \textbf{Repetitions} for repeatedly responding with the same text, 
(2) \textbf{Unrelated} for irrelevant responses to what users request, 
(3) \textbf{Not-aligned} for responses contradicting with previous context, 
and (4) \textbf{Awkward} for grammatically wrong or unfluent responses. 
We also add a free-form answer field where participants can describe the situations that do not fall within the above four categories. 
For each study session, we invite one or two participants to a Zoom
video call, where a moderator briefs the study and instructs participants to complete \textbf{five} conversations with the TOD model. 
During the group session, participants are not allowed to disclose anything related to their conversations with a chatbot. 
The moderator supports participants only when they encounter technical issues. 
When the chatbot provides wrong responses, participants are guided to repeat their original intent up to two times, as we expect the TOD model to recover from its mistakes.
Moreover, participants can continue the conversation with their own arbitrary goals if the chatbot cannot provide services related to the given goals because it is possible for the chatbot to fail to search entities satisfying all requests from users (even in closed-goal settings).

\section{Analysis}
\label{sec:analysis}

\subsection{Dataset and Descriptive Statistics}
We collect 49 open-goal and 50 closed-goal dialogues from 20 participants; due to technical issues, one open-goal participant missed a conversation. The open-goal dialogues consisted of an average 10.53 turns (an adjacent pair of the user and system messages; $SD~\footnote{Standard deviation.}=4.33$), whereas the closed-goal dialogues had 8.92 turns on average ($SD=3.62$). 

\begin{figure}[t!]
    \centering
    \includegraphics[width=\columnwidth]{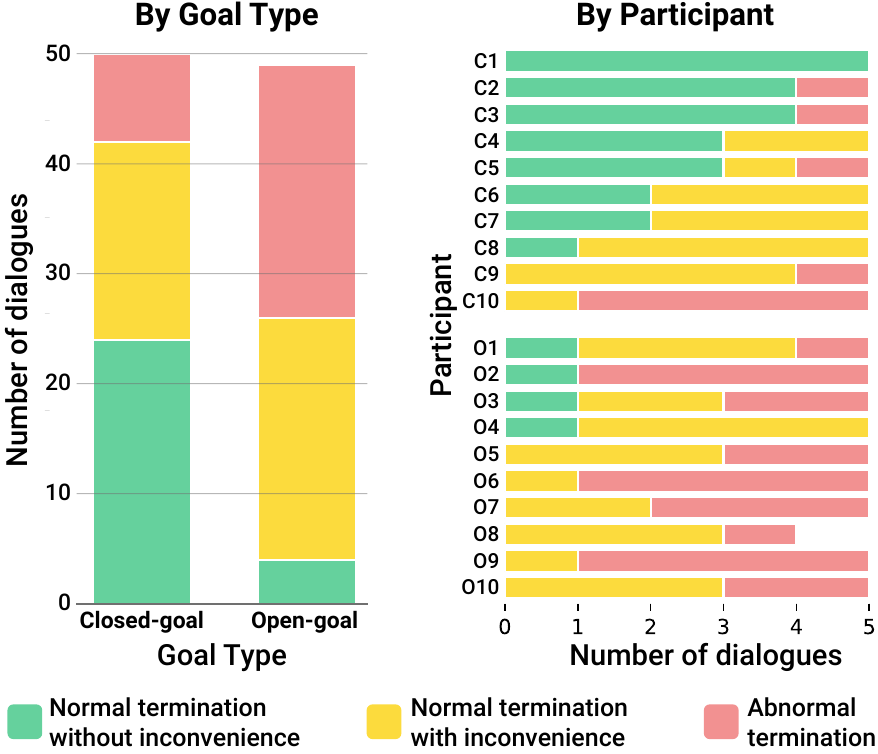}
    \caption{Distributions of the dialogue termination by goal type (left) and participant (right). The green bar refers to situations that users finish the conversation with satisfaction, and the yellow bar refers to situations that users finished the conversation but experience some errors or inconveniences. Lastly, the red bar expresses the proportion of users' strong dissatisfaction by forcedly stopping the conversation. \texttt{C1-C10} and \texttt{O1-O10} denote users with a closed-goal and open-goal, respectively. }
    \label{fig:termination}
    \vspace{-3mm}
\end{figure}
\begin{table*}[t!]
    % \renewcommand{\arraystretch}{1.1}
    % Original arraystretch: 1.2
    \centering\small\sffamily
			\def\arraystretch{1.4}
		    \setlength{\tabcolsep}{0.3em}
    \begin{tabularx}{\textwidth}{|l!{\color{lightgray}\vrule}>{\raggedright}X!{\color{lightgray}\vrule}cc|}
        \hline
        \rowcolor{tableheader}
        \textbf{Error Type} & \textbf{Decription} & \multicolumn{2}{c|}{\textbf{Proportion}} \tabularnewline
        \rowcolor{tableheader} & & Closed & Open \tabularnewline
        \hline
        Irrelevant & Irrelevant responses from the given dialogue context. & 14.6\% & \textbf{23.4}\% \tabularnewline
        \arrayrulecolor{lightgray}\hline
        Self-Contradiction & Contradictory responses with previous bot's responses. & 4.2\% & 5.8\%  \tabularnewline
        \hline
        Repetition & Unnecessary repeated responses with the same semantics.  & 4.8\% & 6.6\%  \tabularnewline
        \hline
        Poor Fluency & Awkward or grammatically broken responses & 4.8\% & 3.7\%  \tabularnewline
        \hline
        Pretending & Hallucinated responses on unverifiable requests & 2.4\% & \textbf{23.6}\%  \tabularnewline
        \hline
        Miscellaneous & All other less frequent errors & 2.9\% & 2.7\%  \tabularnewline
        \arrayrulecolor{black}\hline
    \end{tabularx}
    \caption{Definitions of each error type and corresponding proportion by two goal-types. As demonstrated in the bold text, while other error types occur with the similar proportion, the irrelevancy and pretending problems occur significantly often in the open-goal circumstance (8.8\%p and 21.2\%p more often, respectively).}
    \label{table:error_types}
\end{table*}

\subsection{Dialogue Stability}
Figure \ref{fig:termination} shows the proportion of forced termination during our experiment. 
We find that only 8\% (4 out of 49) of the total open-goal dialogues have finished without any inconveniences, while almost half of the closed-goal dialogues (24 out of 50) show normal termination without any inconveniences. 
Meanwhile, it is important to note that more than half of the dialogues in both goal types had problematic situations for participants.
Statistical tests (and Figure~\ref{fig:termination}) reveal that open-goal settings result in significantly more erroneous dialogues.
We describe the analysis method below, but, in short, we find interactive conversations in the wild to have a clear difference from static benchmark evaluations for both goal settings and especially for the open-goal setting. 

\paragraph{Termination Pattern}
To assess the difference in termination patterns between the two goal types, we use \textit{mixed-effect models}.
These multi-level linear regression models can model the effect of the independent variables (\ie, fixed effect) while controlling the random variance among subjects (\ie, random effect) where multiple data points came from the same subject~\cite{Pinheiro2000MixedEffects}. Treating each dialogue as a data point, we fit a mixed-effect model to the termination type mapped to a numeric scale (0: normal termination, 1: normal termination with inconvenience, and 2: abnormal termination) in increasing order of severity. We put participants as a random effect and the goal type as a fixed effect to see whether the average severity levels of each group are different. 

The maximum-likelihood test reveals that there is a significant random effect of participants ($p<.0001$, $t(17.98)=5.06$), and a significant fixed effect of goal type ($p=.002$, $t(18.14)=3.71$). The estimated mean of the severity scale is 0.68 for closed-goal ($SE=0.13$) and 1.39 for open-goal ($SE=0.14$) with 95\% confidence. This indicates that the severity levels of termination of the two groups are significantly different, and the open-goal dialogues tend to fall in either normal termination with inconvenience or abnormal termination.

\paragraph{Learning Effect}
We also assess whether there is a learning effect on carrying on a successful conversation as participants go through more conversations. We use a mixed-effect model similar to the above analysis but put the order of dialogue per participant (i.e., 0 to 4) as a fixed effect, instead of the goal type. The maximum likelihood test reveals that the order does not have a significant impact on the model ($p=.85$). This implies that participants do not seem to ``learn'' the system's conversation and request capabilities within 5 times of trials.

\subsection{Erroneous Turn Analysis}
\label{sec:error-turnlevel}

In this section, we thoroughly analyze erroneous dialogue turns that occurred in both goal types by categorizing dominant error types.

We qualitatively annotate the erroneous turns. First, all authors of this paper separately annotate a subset (10 dialogues; $\sim$10\%) of the user study data with candidates for the error category. Then they iterate several discussions to generate a codebook of the errors. After the codebook is defined and agreed upon, the main author revisits the entire dataset to annotate erroneous turns. As a result, we identify six types of errors that commonly appeared in the conversations (Table \ref{table:error_types}).
We further describe each error category and corresponding example in the following paragraphs. We denote \symboluser{} as a user and \symbolbot{} as a system in the examples.

\paragraph{Irrelevant}

If the system response is irrelevant to the dialogue context, especially to the previous user utterance, we annotate such responses as \texttt{Irrelevant}. It is the most frequent error type in the user study. Moreover, the error proportion from open-goal dialogues is significantly higher, 8.8\% points, than that from closed-goal dialogues. The following quote is an illustrative example by \texttt{O1}:

\begin{dialogue}
\speak{\symboluser{}} \userdial{Please let me know the \textbf{cancellation policies} of Alexander.}
\speak{\symbolbot{}} \botdial{Yes, I will work on getting this booked for you.}
\end{dialogue}

\paragraph{Self-Contradiction}

\texttt{Self-Contradiction} represents cases that the system response is contradictory with the previous system's responses. This error is severe because it makes the users confused and also causes doubt over the systems in terms of proceeding given task, as illustrated in the below quote by \texttt{C9}. We find that open-goal dialogues contain 1.6\% points more self-contradiction than closed-goal dialogues. 

\begin{dialogue}
\speak{\symboluser{}} \userdial{how about \textbf{3 stars}?}
\speak{\symbolbot{}} \botdial{\textbf{Bridge guest house is available}, would you like to try that?\\ $\cdots$}
\speak{\symboluser{}} \userdial{I want 3 stars guesthouse which has a parking lot.}
\speak{\symbolbot{}} \botdial{\textbf{I don't have anything 3 stars.} would you like to try a different area?\\ $\cdots$}
\speak{\symboluser{}} \userdial{\textbf{Thank you. Can you repeat my reservation?}}
\end{dialogue}

\paragraph{Repetition}

\texttt{Repetition} refers to cases that the system repeats unnecessary responses across multiple turns. This error is also one of the main obstacles to achieving task success as it often annoys the users to quit the conversation. The following quote shows a severe repetition error that happened during our user study (by \texttt{O5}):

\begin{dialogue}
\speak{\symboluser{}} \userdial{Yes. Can you make a reservation for 11:45?}
\speak{\symbolbot{}} $\cdots$ \botdial{Can you tell me the day for the reservation, \textbf{the number of people} in your party, and  \textbf{the number of people}?}
\speak{\symboluser{}} \userdial{Tomorrow. 2 people.}
\speak{\symbolbot{}} \botdial{I have your table booked for 2 people on \textbf{Tuesday} at 11:45, $\cdots$ \textbf{What day} would you like to dine?}
\speak{\symboluser{}} \userdial{You already said Tuesday! Thank you for your help.}
\end{dialogue}

\paragraph{Poor Fluency}

When the system's response is awkward from a given dialogue context or itself, we judge the response as a \texttt{Poor Fluency}. Also, grammatical error belongs to this type. Closed-goal dialogues include about 1\% more \texttt{Poor Fluency} errors than open-goal dialogues. The following example represents this type of error (It is from a conversation with \texttt{C8}).

\begin{dialogue}
\speak{\symboluser{}} \userdial{Hmm, that's not that I want. Can you cancel the current reservation and make one for an African restaurant?}
\speak{\symbolbot{}} \botdial{\textbf{No, it does not.} I can try a new restaurant if you would like.}
\end{dialogue}

\paragraph{Pretending}

\texttt{Pretending} represents system behaviors pretending to handle user's requests beyond the system's scope or answer user's non-verifiable questions. For instance, in the below example by \texttt{O4}, the system provides false information to the users because the presence of a pet-friendliness is not verifiable information for the system. \texttt{Pretending} is the most prominent error type of the open-goal setup; meanwhile, it occurs about 10 times more than in the closed-goal setting. This indicates that this \texttt{Pretending} error might have been overlooked in previous works that focused on static closed-goal evaluations. We further describe and discuss more details of this pretending error type in Section \ref{sec:hallucination}.

\begin{dialogue}
\speak{\symboluser{}} \userdial{Is Wagamama \textbf{pet-friendly}?}
\speak{\symbolbot{}} \botdial{\textbf{Yes it is.} Would you like to book it for you?}
\speak{\symboluser{}} \userdial{Yes. Today at 7pm please.}
\end{dialogue}

\paragraph{Miscellaneous}

The \texttt{Miscellaneous} category, which is less frequent, usually includes arbitrary system behaviors like suddenly making a booking request and ignoring the user's intent. The following example by \texttt{C10} describes such a situation.
 
\begin{dialogue}
\speak{\symboluser{}} \userdial{Then, is the Indian Cuisine you mentioned earlier available for reservation on Wednesday at 15:45?}
\speak{\symbolbot{}} \botdial{Yes it is. \textbf{I have booked it for you} and your reference number is 64o8zpsl.}
\speak{\symboluser{}} \userdial{\textbf{Suddenly?} The person I'm going with just now asks if we can change it to the same time as Thursday?}
\speak{\symbolbot{}} \botdial{Ok. I will try to book that for you.}
\end{dialogue}

\subsection{Participant Feedback}

\begin{figure}[t!]
    \centering
    \includegraphics[width=\columnwidth]{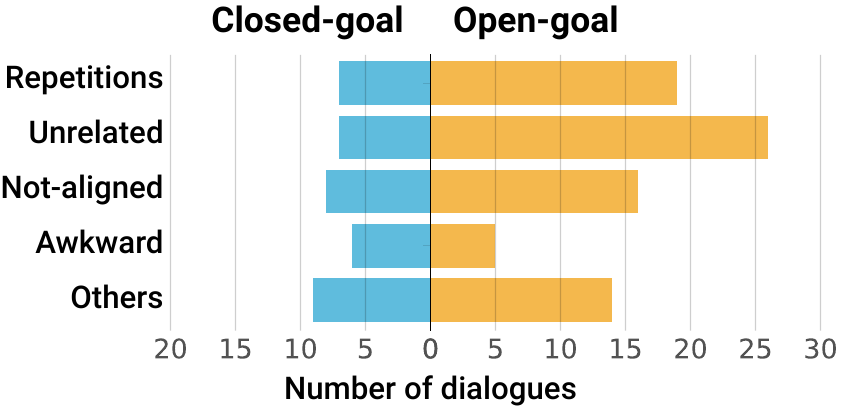}
    \caption{Inconvenience categories collected from participants by two goal types. It can be seen as an implicit or explicit reason to terminate a conversation from the perspective of the user.}
    \label{fig:errors}
\end{figure}
\label{sec:error-diallevel}
We collect negative feedback from participants at a dialogue-level if there are any inconveniences during the conversation. Figure \ref{fig:errors} shows the inconvenience categorization results. As expected, open-goal dialogues show a significantly higher proportion in most inconvenience categories than closed-goal dialogues because the model must have difficulty in handling novel requests in the open-goals.
In addition to the inconvenience categories we specify, we also collect feedback from users in a text form (\texttt{Others} category in Figure \ref{fig:errors}), 
and the full table after removing duplicated feedback will be made public in the Appendix.
% and the full table after removing duplicated feedback can be found in Appendix Table \ref{table:feedback}. 
Interestingly, we find that most inconveniences align with our error categories of Section~\ref{sec:error-turnlevel} well except for the \texttt{Pretending} error type. In other words, the pretending behaviors are \textit{not} identified as inconveniences from the users' perspective.

\subsection{Why does \texttt{Pretending} Matter?}
\label{sec:hallucination}

In this subsection, we take a deeper look at the ``pretending'' behaviors of the TOD systems. The pretending behavior is similar to the hallucination problem~\citep{ji-et-al-2022-hallucination} prevalent in generative models in that it provides false information to the users. However, one distinct feature of the pretending behavior is that it is \textbf{hard to be recognized as an error at the time of occurrence} by only its surface form because most TOD systems rely on variable service-specific knowledge that users cannot easily access while using the service. It also differs from the knowledge base inconsistency (KBI) in \citet{qin-etal-2021-dont}. While the KBI only regards the wrong responses based on ``verifiable'' knowledge, \texttt{Pretending} indicates responses over ``non-verifiable'' knowledge beyond the system's scope.

It is a severe problem for both agents and users since it interrupts accurate decision-making to achieve users' goals.
For example, other error types such as \texttt{Irrelevant}, \texttt{Self-Contradiction}, and \texttt{Repitition} can be easily recognized as superficial problems by the users. In those cases, the users can avoid unwanted conversation flow by complaining and terminating the conversations. However, when the TOD system naturally responds to users' exceptional requests and does not take corresponding action behind, users have no way to perceive the fact that the ongoing conversation is wrong from their initial requests. For this reason, the \texttt{Pretending} is not exposed in \textit{any} user evaluation shown in Figure \ref{fig:termination}, Figure \ref{fig:errors}, or Appendix Table \ref{table:feedback}. 
In other words, even users who normally terminated dialogue without any inconvenience (\ie, green bar in Figure \ref{fig:termination}) can suffer from the pretended dialogues.

\subsection{\update{Can Large Language Models Solve TOD Problems?}}

\update{Although previous studies demonstrate the imperfectness of large language models in TOD systems~\cite{jakobovits-et-al-2022-conversational, bang-etal-2023-multitask}, we conduct case studies to verify large language models' capacities to resolve aforementioned conversational errors. Since it is impossible to equalize every experimental setup between fine-tuned language models and proprietary large language models\footnote{\texttt{gpt-4-turbo} in our case studies.}, we proxy intermediate modules of TOD systems with instruction prompts by modifying those of \citet{chung-etal-2023-instructtods}. On top of that, we pinpoint erroneous turns in Section \ref{sec:error-turnlevel} and compare generated responses of each model. We describe detailed setup such as prompts and action definition in Section \ref{sec:llm_setup}.}

\update{Based on the result of our case studies, as illustrated in Table \ref{table:llm_resp}, we observe that large language models  mitigate most conversational inconvenience, presumably due to their strong zero/few-shot capabilities on unseen domains. However, regarding \texttt{Pretending}, responses of both cases contain significant flaws. Specifically, according to Table \ref{table:llm_resp}, both conventional TOD model and large language model do not recognize the fact that pet-friendliness does not belong to the service range and provide untrustworthy responses, which can lead to physical harm (\eg, wrong reservation) in real services. One of the expected causes is language models' overconfidence of unseen scenarios, but we also find that predefined actions given to TOD models are confined to deal with diverse situations in a flexible manner\footnote{Notably, there are roughly-defined actions in conventional TOD scenarios, such as \texttt{inform, request, recommend, etc.}}. Controlling overconfidence in language models~\citep{miao-etal-2021-prevent, mielke-etal-2022-reducing} can partially resolve conversational errors, but defining available actions mostly belongs to the range of system design, especially in service-specific scenarios. We further discuss the future direction in Section \ref{sec:discussion}.}
\section{Conclusion}
In this work, we demonstrate user familiarity bias in current TOD benchmarks, which the recent TOD research community has overlooked.
To effectively unveil the bias, we contrast two user groups with different user goals via an interactive user study. Against the closed-goal within the constrained scenarios, we introduce a control user group by assigning unconstrained scenarios to the participants, namely open-goal. Users in the two groups converse with the academically-discussed TOD chatbot following the given closed or open-goals. Our study reveals the TOD system exposed to the user familiarity bias significantly fails to converse with the users with open-goals. We identify prevalent error types by analyzing the resulting conversations. Furthermore, we highlight the pretending behaviors of the TOD system with its characteristics and toxicity, \update{which are not easily solved by simply utilizing large language models.}
\section{Limitations}
Regarding the participants of our user study, all of them are internal employees of a giant tech company, the majority of whom are highly educated (60\% hold a master's or doctoral degree). However, they show various experiences with chatbots, not correlated with educational degree. Some of them are bilingual, while others are not native English speakers. Furthermore, since we assume a traveling situation, the conversational scenario was not challenging, even for non-native speakers. 

\section{Ethics Statement}
In our user study, we collected demographic information such as name, age, gender, the highest level of education, occupation, native language, and experiences with the AI chatbot, after informing them that it would be used only for research purposes and acquiring their consent. \update{We clearly introduced the purpose of our study and the usage of collected information before experiments, and all participants consented to our instructions.}

Throughout the interaction with the chatbot, we instructed participants to play the role of potential users only, without disclosing any personally identifiable information about themselves. Collected dialogues were de-identified by giving anonymized user IDs. 
Throughout the annotating process, the authors examined all the gathered conversations, and no offensive content was found. Participants took part in the chat for roughly 30 minutes and were compensated with a 5,000 KRW (equivalent to 3.7 USD) gift card, which was somewhat higher than the Korean minimum wage at that time.

\bibliography{anthology,custom}
\bibliographystyle{acl_natbib}

\clearpage

\appendix
\newpage \onecolumn

\section{Participant Information}
\label{appendix:participant}
Participants are aged between 24 to 35 ($Median=29.5$), and 12 are female. 
Three participants report that they are native English speakers. 
Eight participants have used AI chatbots for less than one month. 
Participants consist of 7 infrastructure engineers, 6 software engineers, 3 AI research scientists, 2 self-employed, a UX designer, and a data scientist. Detailed information can be found in Table \ref{table:demographics}.

\begin{table*}[h!]
    \centering\small\sffamily
			\def\arraystretch{1.5}
		    \setlength{\tabcolsep}{0.5em}
    % \newcolumntype{Y}{>{\centering\arraybackslash\hsize=0.3\hsize}X}
    \begin{tabularx}{\textwidth}{|c!{\color{lightgray}\vrule}c!{\color{lightgray}\vrule}c!{\color{lightgray}\vrule}c!{\color{lightgray}\vrule}c!{\color{lightgray}\vrule}>{\centering\arraybackslash}X|}
        \hline
        \rowcolor{tableheader}
        \textbf{User ID} & \textbf{Age Range} & \textbf{Gender} & \textbf{Education} & \textbf{Occupation} & \textbf{Chatbot Proficiency} \tabularnewline
        \hline
        C1 & 26-30 & Male & Bachelor's & Infrastructure Engineer & Less than 1 month \tabularnewline
        \arrayrulecolor{lightgray}\hline
        C2 & 26-30 & Female & Master's & Software Engineer & Less than 1 month \tabularnewline
        \hline
        C3 & 26-30 & Male & Master's & Software Engineer & Less than 1 month \tabularnewline
        \hline
        C4 & 26-30 & Male & Master's & Data scientist & 1 year to 3 year \tabularnewline
        \hline
        C5 & 26-30 & Female & Bachelor's & Software Engineer & 1 year to 3 year \tabularnewline
        \hline
        C6 & 26-30 & Female & Bachelor's & Software Engineer & Less than 1 month \tabularnewline
        \hline
        C7 & 31-35 & Male & Master's & UX Designer & 1 year to 3 year \tabularnewline
        \hline
        C8 & 31-35 & Female & Master's & Research Scientist & 1 year to 3 year \tabularnewline
        \hline
        C9 & 31-35 & Male & Ph.D./M.D. & Infrastructure Engineer & Less than 6 months \tabularnewline
        \hline
        C10 & 31-35 & Female & Bachelor's & Self-employed & 1 year to 3 year \tabularnewline
        \arrayrulecolor{black}\hline
        
        O1 & 26-30 & Male & Ph.D./M.D. & Infrastructure Engineer & Less than 1 month \tabularnewline
        \arrayrulecolor{lightgray}\hline
        O2 & 21-25 & Female & Bachelor's & Software Engineer & Less than 1 month \tabularnewline
        \hline
        O3 & 21-25 & Female & Bachelor's & Infrastructure Engineer & Less than 1 year \tabularnewline
        \hline
        O4 & 26-30 & Female & Bachelor's & Self-employed & 1 year to 3 year \tabularnewline
        \hline
        O5 & 26-30 & Female & Master's & Infrastructure Engineer & Less than 1 month \tabularnewline
        \hline
        O6 & 21-25 & Female & Bachelor's & Infrastructure Engineer & Less than 1 month \tabularnewline
        \hline
        O7 & 21-25 & Female & Master's & Infrastructure Engineer & Less than 6 months \tabularnewline
        \hline
        O8 & 31-35 & Male & Ph.D./M.D. & Research Scientist & More than 3 years \tabularnewline
        \hline
        O9 & 31-35 & Male & Master's & Software Engineer & 1 year to 3 year \tabularnewline
        \hline
        O10 & 31-35 & Female & Ph.D./M.D. & Research Scientist & More than 3 years \tabularnewline
        \arrayrulecolor{black}\hline
    \end{tabularx}
    \caption{Participant information of our user study. We anonymize the name of each participant by assigning user ID and categorizing the range of their age. Users whose ID starts with \texttt{C} conduct closed-goal conversation, whereas those whose ID starts with \texttt{O} conduct open-goal conversation.
    }
    \label{table:demographics}
\end{table*}

\section{Model Implementation Details}
For the TOD system in our experiments, we use the public implementation of GALAXY\footnote{\url{https://github.com/siat-nlp/GALAXY}. Apache license 2.0.}~\citep{he2022galaxy}. The model specification follows \citet{he2022galaxy}; initialized with UniLM~\citep{dong2019unified}, which has a transformer-based architecture with 109M parameter size. We fine-tune this model on MultiWOZ 2.1\footnote{\url{https://github.com/budzianowski/multiwoz}. MIT license.}~\citep{eric-etal-2020-multiwoz}. We follow the default hyper-parameter settings provided by the authors. Training is completed within a few hours using 1 NVIDIA A100. PyTorch\footnote{\url{https://pytorch.org/}} library is used for model training, and NLTK\footnote{\url{https://www.nltk.org/}} and spaCy\footnote{\url{https://spacy.io/}} are for text processing. We implemented a chat interface on Gradio\footnote{\url{https://gradio.app/}}~\citep{abid2019gradio} web platform. At inference time, greedy search is used for output prediction.

\section{\update{Case Studies on Large Language Model}}
\label{sec:llm}

\subsection{\update{Setup}}
\label{sec:llm_setup}

\update{With a slight modification of \citet{chung-etal-2023-instructtods}, we build prompts for action decision and response generation, demonstrated in Table \ref{table:prompt_llm}, in order to simulate TOD systems utilizing large language models. We adopt predefined actions in GALAXY model as available actions. For end-to-end simulation, large language models generate responses by referring to conversation context and selected action (which is also generated).}

\begin{table}[h!]
    % Original arraystretch: 1.4
    \centering\small\ttfamily
			\def\arraystretch{1.4}
		    \setlength{\tabcolsep}{0.5em}
    \begin{tabularx}{\columnwidth}{|>{\raggedright}X|}
     \hline
     \rowcolor{tableheader}\textbf{\textsf{Prompt for Action Decision}} \tabularnewline
     \arrayrulecolor{lightgray}\hline
     \small
    In a task-oriented dialogue setting, generate an appropriate system ACT to the USER query in the conversation provided in CONTEXT. A single system ACT should be selected within the list of available actions. Make sure that selected action should not contradict with previous conversation. Only generate the selected action.\newline
    
    You can follow this example:\newline\newline
    Conversation domain: restaurant\newline
    Available actions: ["inform", "request", "nooffer", "recommend", "select", "offerbook"]\newline
    CONTEXT:\newline
    USER: I need a place to eat foods.\newline
    SYSTEM: Any preference in the type of food?\newline
    USER: Preferably Italian, but Chinese is also fine.\newline 
    ACT: recommend \newline

    Conversation domain: \{\{domain\}\}\newline
    Available actions: \{\{action\}\}\newline
    CONTEXT:\newline
    \{\{context\}\}\newline
    ACT: 
     \tabularnewline
     \arrayrulecolor{black}\hline
     \rowcolor{tableheader}\textbf{\textsf{Prompt for Response Generation}} \tabularnewline
     \arrayrulecolor{lightgray}\hline
     \small
    In a task-oriented dialogue setting, generate a natural and helpful SYSTEM response to the USER query in the conversation provided in CONTEXT. You should follow the information provided in ACT to generate this answer. Do not mention that you are referring to a dataframe and don't overload the user with too many choices.\newline

    You can follow this example:\newline\newline
    USER: I need a place to eat foods.\newline
    SYSTEM: Any preference in the type of food?\newline
    USER: Preferably Italian, but Chinese is also fine.\newline
    ACT: [restaurant] recommend\newline
    SYSTEM: How about in Bouly's Italy?\newline\newline
    
    CONTEXT:\newline
    \{\{context\}\}\newline
    ACT: \{\{action\}\}\newline
    SYSTEM:  \tabularnewline
     \arrayrulecolor{black}\hline
    \end{tabularx}
    \caption{Prompts used for large language model simulation.}
    \label{table:prompt_llm}
    \vspace{-3mm}
\end{table}

\subsection{\update{Response Comparison}}
\label{sec:llm_result}
\update{We compare responses of GALAXY model with those of proprietary large language model on sampled dialogues, specifically mentioned in Section \ref{sec:error-turnlevel}. Table \ref{table:llm_resp} demonstrates each example with partial dialogue turns.}

\begin{table*}[t!]
    \centering\small\sffamily
			\def\arraystretch{1.2}
		    \setlength{\tabcolsep}{0.4em}
    % \newcolumntype{Y}{>{\centering\arraybackslash\hsize=0.3\hsize}X}
    \begin{tabularx}{\textwidth}{|c!{\color{lightgray}\vrule}X!{\color{lightgray}\vrule}X|}
        \hline
        \rowcolor{tableheader}
        \textbf{Error Type} & \textbf{Response (GALAXY)} & \textbf{Response (\texttt{gpt-4-turbo})} \tabularnewline
        \hline

        % Irrelevant
        \multirow{8}{*}{Irrelevant} & \multicolumn{2}{X|}{\vspace{-2mm}\shortstack[l]{
            \symboluser{}  No. tell me the other option. \\
            \symbolbot{} The alexander is located at 56 saint barnabas road and the phone number \\is 01223525725. \\
            \symboluser{} Please let me know the {\color{blue}cancellation policies} of Alexander.
        }
        } 
        \tabularnewline\arrayrulecolor{lightgray}\cline{2-3} 
        & \multirow{4}{*}{\shortstack[l]{\symbolbot{} Yes, I will work on {\color{red}getting this booked}\\for you.}} 
        & \multirow{4}{*}{\shortstack[l]{\symbolbot{} I apologize, but I don't have information \\about {\color{blue}cancellation policies} for the Alexander \\Bed and Breakfast. Would you like me to \\help you with anything else?}} 
        \newline\newline\newline\newline \tabularnewline \arrayrulecolor{lightgray}\hline \arrayrulecolor{black}

        % Self-Contradiction
        \multirow{12}{*}{Self-Contradiction} & \multicolumn{2}{X|}{\vspace{-2mm}\shortstack[l]{
            \symboluser{}  How about 3 stars? \\
            \symbolbot{} {\color{blue}Bridge guest house is available}, would you like to try that ? \\
            \symboluser{} Okay. I have a car. Is there a parking lot? \\
            \symbolbot{} I am sorry , there isn't anything that meets your criteria . \\
            \symboluser{} I want 3 stars guesthouse in the south which has a parking lot. 
        }
        } 
        \tabularnewline\arrayrulecolor{lightgray}\cline{2-3} 
        & \multirow{5}{*}{\shortstack[l]{\symbolbot{} {\color{red}I don’t have anything 3 stars.} \\would you like to try a different area?}} 
        & \multirow{5}{*}{\shortstack[l]{\symbolbot{} I'm sorry, but currently, there are {\color{blue}no}\\ {\color{blue}3-star guesthouses in the south area that}\\ {\color{blue}offer parking facilities.} Is there anything else \\I might assist you with or perhaps another\\ area you would consider?}} 
        \newline\newline\newline\newline\newline \tabularnewline \arrayrulecolor{lightgray}\hline \arrayrulecolor{black}

        % Repetition
        \multirow{10}{*}{Repetition} & \multicolumn{2}{X|}{\vspace{-2mm}\shortstack[l]{
            \symboluser{}  Yes. Can you make a reservation for 11:45? \\
            \symbolbot{} I would be happy to. can you tell me {\color{blue}the day for the reservation}, the number of people \\in your party, and the number of people. \\
            \symboluser{} Tomorrow. 2 people. 
        }
        } 
        \tabularnewline\arrayrulecolor{lightgray}\cline{2-3} 
        & \multirow{5}{*}{\shortstack[l]{\symbolbot{} I have your table booked for 2 people on \\Tuesday at 11:45 , your reference number is \\zbznwq27. {\color{red}What day would you like to dine?}}} 
        & \multirow{5}{*}{\shortstack[l]{\symbolbot{} Great! Your reservation at Curry Prince \\{\color{blue}for 2 people tomorrow at 11:45 AM is all set.} \\If you need to make any changes or have\\ more requests, feel free to let me know. \\Enjoy your meal!}} 
        \newline\newline\newline\newline\newline \tabularnewline \arrayrulecolor{lightgray}\hline \arrayrulecolor{black}
        
        % Poor Fluency
        \multirow{11}{*}{Poor Fluency} & \multicolumn{2}{X|}{\vspace{-2mm}\shortstack[l]{
            \symboluser{}  Does the place offer halal or African food? \\
            \symbolbot{} It serves modern european food.\\
            \symboluser{}  Hmm, that's not that I want. Can you {\color{blue}cancel the current reservation} and make one for \\an African restaurant?
        }
        } 
        \tabularnewline\arrayrulecolor{lightgray}\cline{2-3} 
        & \multirow{6}{*}{\shortstack[l]{\symbolbot{} {\color{red}No, it does not.} I can try a new \\ restaurant if you would like.}} 
        & \multirow{6}{*}{\shortstack[l]{\symbolbot{} I've {\color{blue}cancelled your reservation} at the\\ Cambridge Chop House. I can book a table \\for you at Bedouin, which serves African \\cuisine. Would you like me to proceed with \\the booking for 6 people this Monday at \\17:15?}} 
        \newline\newline\newline\newline\newline\newline \tabularnewline \arrayrulecolor{lightgray}\hline \arrayrulecolor{black}

        % Pretending
        \multirow{8}{*}{Pretending} & \multicolumn{2}{X|}{\vspace{-2mm}\shortstack[l]{
            \symboluser{}  It's just going to be me. Make sure its pet-friendly though. \\
            \symbolbot{} It is expensive. What time would you like to dine? \\
            \symboluser{} Is Wagamama {\color{blue}pet-friendly}?
        }
        } 
        \tabularnewline\arrayrulecolor{lightgray}\cline{2-3} 
        & \multirow{4}{*}{\shortstack[l]{\symbolbot{} {\color{red}Yes it is.} Would you like to book it for you?}} 
        & \multirow{4}{*}{\shortstack[l]{\symbolbot{} Wagamama {\color{red}does not allow pets inside.}\\ Would you like to find another restaurant or \\something else?}} 
        \newline\newline\newline\newline \tabularnewline \arrayrulecolor{black}\hline 
        
    \end{tabularx}
    \caption{\update{Comparison of GALAXY and \texttt{gpt-4-turbo} (LLM) on sampled dialogues in our user studies. Upper part of conversation on each error type illustrates partial context. For large language model prompt, full dialogues become an input as a context. \textbf{Irrelevant:} While GALAXY does not handle cancellation policies, LLM correctly understands the requirement and provides relevant response. \textbf{Self-Contradiction:} By providing explicit context (\ie, no 3-star guesthouses with parking), response of LLM does not contradict with previous conversation. \textbf{Repetition:} LLM mostly does not unnecessarily ask the same question. \textbf{Poor Fluency: } LLM shows fluent conversational skills by understanding the current topic of the conversation. \textbf{Pretending: } Both GALAXY and LLM have no capability to verify whether Wagamama is enrolled as a pet-friendly restaurant. Therefore, both responses are not true.}}
    \label{table:llm_resp}
\end{table*}

\clearpage
\section{Discussion}
\label{sec:discussion}
As the importance of real user study or interactive evaluation gets bigger, we discuss the future direction in two folds: enhancing transparency to let users trust the chatbot, and managing fallback situations to detect users' exceptional requests.

\subsection{Enhancing Transparency}
Our participants often struggle to confirm their booking options as in Appendix Table~\ref{table:feedback}. 
As a way to improve the reliability of TOD systems, we suggest enhancing \textit{transparency} of the system, which has been actively discussed in the HCI community~\cite{saleema2019guidelines}. 
Transparency is a mechanism exposing hidden (\ie, non-obvious) information to users who have difficulty in getting the information directly~\cite{emilee2018explanations}. As our findings show that the lack of \textit{user familiarity} provokes various inconveniences including the pretending problem, TOD systems in natural language processing field should also be designed to display intermediate by-products during the conversation in order to provide explainable rationales for their decisions~\cite{saleema2019guidelines, vera2020questioning}.

In the era of billion-scale large language models, the necessity of transparency is still valid. Although emerging works on grounded LLMs (\textit{e.g.}, \citet{yao2023react}, ChatGPT with plugins~\citep{openai2023gpt4} try to enhance trustworthiness using executable sources, they are still exposed to familiarity bias problem as long as they keep black-boxed service pipeline. 

\subsection{Managing Fallback Situation}
Users with low familiarity with the system inevitably make exceptional requests. As we can find in user comments in Appendix Table \ref{table:feedback}, 
a large number of users in an open-goal setup go through irrelevant and pretending responses from the bot. We emphasize the need to recognize exceptional requests and manage fallback situations towards robust TOD systems.

\paragraph{Out-of-Scope Detection}
 In the field of intent classification, previous literature has studied detecting out-of-scope intents to prevent generating erroneous responses from the original intent \citep{larson-etal-2019-evaluation, zhang-etal-2022-pre-trained, zhang-etal-2022-new, cho2022enhancing}. Moreover, \citet{shrivastava-etal-2021-saying} try to generate contextualized fallback responses to users' exceptional requests. 
 However, more datasets for fallback detection are required especially for multi-turn and multi-domain TOD scenarios beyond the single-turn detection scenarios.

\paragraph{Handling Request as Unstructured Form}
\citet{kim-etal-2020-beyond} combine unstructured knowledge, FAQ pairs, with structured knowledge, DBs. The work includes (unstructured) knowledge-seeking turn detection to handle domain-specific requests with FAQs beyond the scope of structured knowledge. However, the work still assumes high user familiarity, \ie, it always contains relevant knowledge for a given request. We believe retrieval-augmented detection leveraging the FAQ pairs can be a promising approach to strengthen the approach towards a low user familiarity setup effectively~\cite{thulke2021efficient}.

On the other hand, typical dialogue state tracking to access structured knowledge is not robust in terms of handling exceptional requests since it works based on \textit{predefined} slots. \citet{bae2022keepmeupdated} adopts a text-formed dialogue state by summarising the dialogue context for effective memory management in multi-session open-domain dialogue. We believe that dialogue management based on unstructured information can have advantages not only in avoiding exceptional requests but also in leveraging advanced language understanding abilities of recent language models at a scale, as its generalizable text format.

\clearpage
\section{Subjective User Feedback}
\begin{table*}[h!]
\small
    			\def\arraystretch{1.2}
    		    \setlength{\tabcolsep}{0.5em}
    		    \centering
\begin{tabular}{|c!{\color{lightgray}\vrule}l!{\color{lightgray}\vrule}l|}
\hline
\rowcolor{tableheader}\textbf{Goal Type} & \textbf{Error Type} & \textbf{Feedback} \\ \hline
\multirow{14}{*}{\textbf{Closed}} & \multirow{3}{*}{\textbf{Relevancy}} & \textit{I told about the reservation conditions, but the chatbot answered irrelevantly.} \\ \arrayrulecolor{lightgray}\cline{3-3} 
 &  & \begin{tabular}[c]{@{}l@{}}\textit{There was an answer that seemed to have forgotten the context of the past,} \\ \textit{but generally the conversation ended without any problems.}\end{tabular} \\ \cline{3-3} 
 &  & \begin{tabular}[c]{@{}l@{}}\textit{I was asked how many people would visit, so I said I was alone. But the chatbot said} \\ \textit{it didn't have a room, and it couldn't continue conversation after that.}\end{tabular} \\ \arrayrulecolor{black}\cline{2-3} 
 & \textbf{Awkward} & \textit{The chatbot said ``Yes, I can'' when I asked the parking availability in the hotel.} \\ \cline{2-3} 
 & \textbf{Repetition} & \textit{The chatbot repetitively asked ``What area would you like to stay in?''.} \\ \cline{2-3} 
 & \textbf{Contradiction} & \begin{tabular}[c]{@{}l@{}}\textit{The chatbot said there is no place to park, but it reversed its saying.}\\\textit{It also told there is a 3-star hotel, then reversed.}\end{tabular} \\ \cline{2-3} 
 & \textbf{Redundant} & \textit{I told the model that I can look up the address by myself, but it gave me the address.} \\ \cline{2-3} 
 & \multirow{7}{*}{\textbf{Booking}} & \textit{The chatbot just ended conversation by just recommending, not booking.} \\ \arrayrulecolor{lightgray}\cline{3-3} 
 &  & \textit{I asked the chatbot to recommend, but it arbitrarily booked it.} \\ \cline{3-3} 
 &  & \textit{It booked without any options I prefer.} \\ \cline{3-3} 
 &  & \textit{It did not confirm my requests.} \\ \cline{3-3} 
 &  & \textit{I wanted to confirm that my reservation is at 9:30 but chatbot did not say.} \\ \cline{3-3} 
 &  & \textit{I wished to reserve Varsity, but the chatbot booked Bloomsbury and did not fix.} \\ \cline{3-3} 
 &  & \textit{I requested to confirm my reservation because I did not trust, but it could not.} \\ \arrayrulecolor{black}\hline
\multirow{27}{*}{\textbf{Open}} & \multirow{16}{*}{\textbf{Relevancy}} & \textit{The bot couldn't understand my additional requests.} \\ \arrayrulecolor{lightgray}\cline{3-3} 
 &  & \textit{The bot couldn't understand and answer my question about additional information.} \\ \cline{3-3} 
 &  & \begin{tabular}[c]{@{}l@{}}\textit{After being asked whether the Asian restaurant serves Italian wines,} \\ \textit{it keeps answering that the Asian restaurant serves Italian food.} \\ \textit{Following correction questions did not work.}\end{tabular} \\ \cline{3-3} 
 &  & \textit{The model does not understand the question correctly.} \\ \cline{3-3} 
 &  & \textit{It does not get back with the list of menus from the pizza hut city centre.} \\ \cline{3-3} 
 &  & \textit{The model keeps saying about night clubs information instead of accommodation.} \\ \cline{3-3} 
 &  & \textit{The chatbot doesn't understand additional requests on gluten-free and pet-friendliness.} \\ \cline{3-3} 
 &  & \begin{tabular}[c]{@{}l@{}}\textit{The chatbot understood ``Slightly more expensive {[}than cheap{]}'' expression as ``expensive,''} \\ \textit{which is wrong.}\end{tabular} \\ \cline{3-3} 
 &  & \begin{tabular}[c]{@{}l@{}}\textit{The chatbot asked whether I wish for a different cuisine,} \\ \textit{when I never stated any in the firstplace.}\end{tabular} \\ \cline{3-3} 
 &  & \textit{I asked for hotel amenities, but the chatbot thought I was asking for the address.} \\ \cline{3-3} 
 &  & \begin{tabular}[c]{@{}l@{}}\textit{I asked whether a certain restaurant serves gluten-free,} \\ \textit{but the chatbot didn't directly address the request.}\end{tabular} \\ \cline{3-3} 
 &  & \textit{It would not answer my question.} \\ \cline{3-3} 
 &  & \textit{It suddenly says ``Your booking was successful , the reference number is i23gx1yf''.} \\ \cline{3-3} 
 &  & \textit{I don't feel like the model remembers the conversation context.} \\ \cline{3-3} 
 &  & \textit{It often made weird responses.} \\ \cline{3-3} 
 &  & \textit{I stopped conversation because it never answer what I asked.} \\ \arrayrulecolor{black}\cline{2-3} 
 & \textbf{Awkward} & \textit{Sometimes the bot would repeat the same options twice in the same sentence.} \\ \cline{2-3} 
 & \multirow{3}{*}{\textbf{Repetition}} & \textit{It made a reservation for tuesday, but still asked me what day I'd like to dine.} \\ \arrayrulecolor{lightgray}\cline{3-3} 
 &  & \textit{Although I answered, it would ask me the same thing again.} \\ \cline{3-3} 
 &  & \textit{Regardless of my answer it just repeats the same thing.} \\ \arrayrulecolor{black}\cline{2-3} 
 & \textbf{Contradiction} & \begin{tabular}[c]{@{}l@{}}\textit{``Since there are several hotel-s in the centre of town we have only 2 guest house.''} \\ \textit{didn't make sense.}\end{tabular} \\ \cline{2-3} 
 & \multirow{2}{*}{\textbf{Redundant}} & \begin{tabular}[c]{@{}l@{}}\textit{Right after booking a guest house,} \\ \textit{the model asked about hotel booking which is unnecessary.}\end{tabular} \\ \arrayrulecolor{lightgray}\cline{3-3} 
 &  & \textit{The chatbot said relevant, but unnecessary questions.} \\ \arrayrulecolor{black}\cline{2-3} 
 & \multirow{4}{*}{\textbf{Booking}} & \textit{I am not sure the chatbot truly understand my booking requests.} \\ \arrayrulecolor{lightgray}\cline{3-3} 
 &  & \begin{tabular}[c]{@{}l@{}}\textit{The chatbot unnecessarily tried to push me into booking the places/restaurants} \\ \textit{when my goal is to simply get information.}\end{tabular} \\ \cline{3-3} 
 &  & \textit{I am not sure if the model really booked successfully.} \\ \cline{3-3} 
 &  & \textit{The chatbot seems to be obsessed with the purpose of booking something.} \\ \arrayrulecolor{black}\hline
\end{tabular}
\caption{A dialogue-level user feedback based on goal types. Similar feedbacks are categorized with error types.}
\label{table:feedback}
\end{table*}

\clearpage
\section{Ontology Used for Open-Goal Dialogues}
\colorlet{punct}{red!60!black}
\definecolor{background}{HTML}{EEEEEE}
\definecolor{delim}{RGB}{20,105,176}
\colorlet{numb}{magenta!60!black}

\lstdefinelanguage{json}{
    basicstyle=\footnotesize\ttfamily,
    numbers=left,
    numberstyle=\scriptsize,
    stepnumber=1,
    numbersep=8pt,
    showstringspaces=false,
    breaklines=true,
    captionpos=b,                    
    frame=lines,
    backgroundcolor=\color{background},
    literate=
     *{0}{{{\color{numb}0}}}{1}
      {1}{{{\color{numb}1}}}{1}
      {2}{{{\color{numb}2}}}{1}
      {3}{{{\color{numb}3}}}{1}
      {4}{{{\color{numb}4}}}{1}
      {5}{{{\color{numb}5}}}{1}
      {6}{{{\color{numb}6}}}{1}
      {7}{{{\color{numb}7}}}{1}
      {8}{{{\color{numb}8}}}{1}
      {9}{{{\color{numb}9}}}{1}
      {:}{{{\color{punct}{:}}}}{1}
      {,}{{{\color{punct}{,}}}}{1}
      {\{}{{{\color{delim}{\{}}}}{1}
      {\}}{{{\color{delim}{\}}}}}{1}
      {[}{{{\color{delim}{[}}}}{1}
      {]}{{{\color{delim}{]}}}}{1},
}

\begin{figure*}[h!]
\begin{lstlisting}[language=json,firstnumber=1,
label={list:ontology}, caption=A postprocessed ontology for exceptional requests generated from GPT-3. We collect raw task descriptions and categorize common scenarios which are not seen during the model training.]
{
    "hotel": {
        "Requests": [
            "Outdoor seating",
            "Rooms with exceptional views",
            "24-hour front desk",
            "Breakfast options",
            "Check-in/out policies",
            "Cancellation policies",
            "Cost around $150 per nights"
        ],
        "Facilities": [
            "Facilities: Gym", 
            "Facilities: Spa", 
            "Facilities: Swimming pool", 
            "Facilities: Outdoor terrace", 
            "Facilities: Non-smoking room"
        ],
        "Amenities": [
            "Amenities: Mini-fridge", 
            "Amenities: Tea and coffee facilities", 
            "Amenities: Private bath"
        ]
    },
    "restaurant": {
        "Requests": [
            "Cost under $50 per person",
            "Open until late",
            "Italian wine serving",
            "List of menus",
            "Comfortable seating with enough space",
            "Delivery or take-out service"
        ],
        "FoodOptions": [
            "Food option: Gluten-free", 
            "Food option: Vegetarian"
        ]
    },
    "common": {
        "Requests": [
            "Wheelchair access",
            "Weather forecast",
            "Website link",
            "Good customer review",
            "Located in safe environment",
            "Accept payment via cards",
            "Easy access to public transportation",
            "Near the tourist attraction"
        ],
        "Friendly": [
            "Pet-friendly", "Kid-friendly", "Family-friendly"
        ]
    }
}
\end{lstlisting} 
\end{figure*}

\clearpage
\section{Demo Interface for User Study}
\begin{figure*}[!ht]
    \centering
    \includegraphics[width=0.8\textwidth]{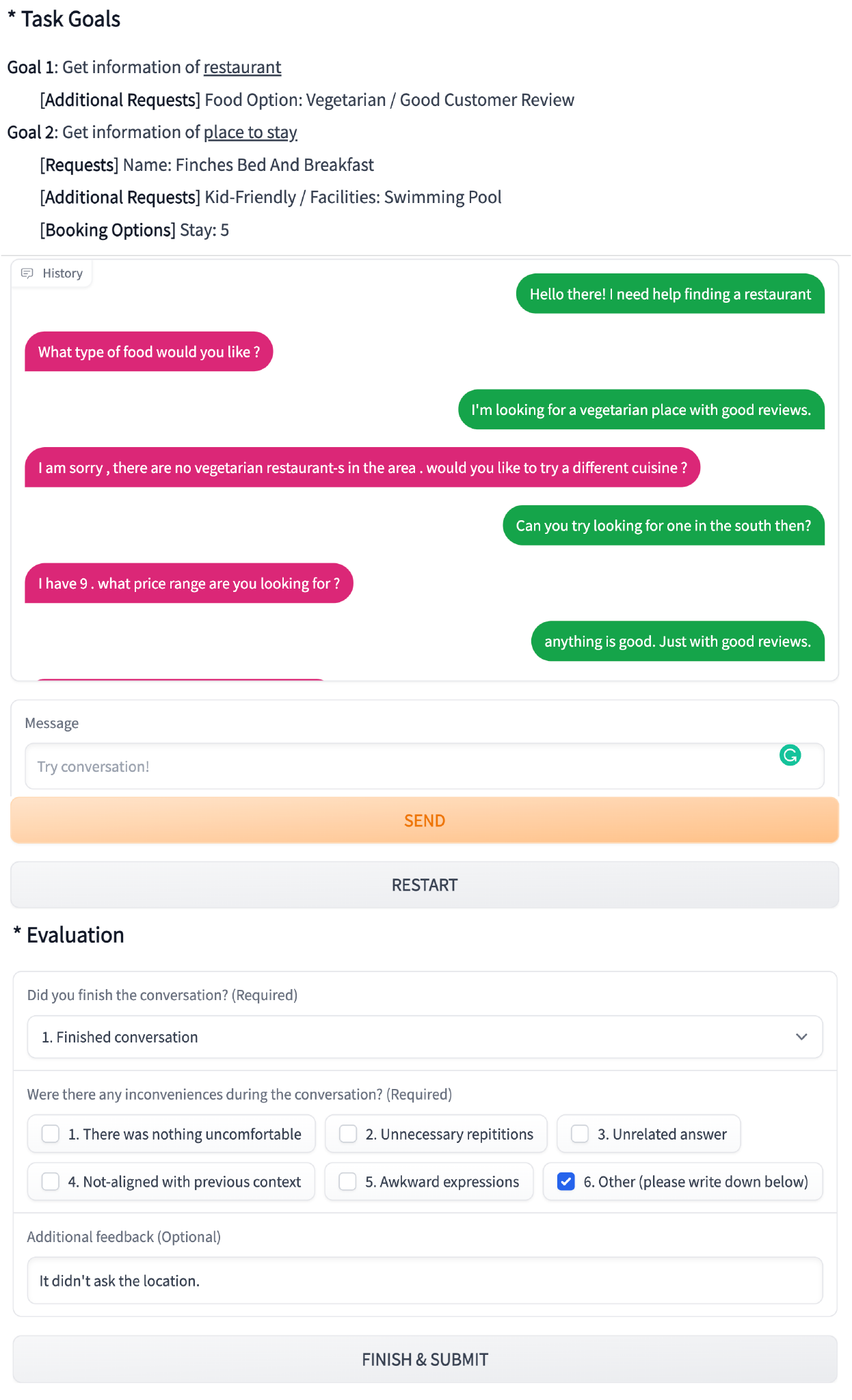}
    \caption{A Gradio demo example of an interactive user study. Unlike sentence-based user goal guidance of conventional Wizard-of-Oz setup, we provide user goals in a structured form.} 
    \label{fig:demo}
\end{figure*}

\end{document}